\documentclass[review]{elsarticle}

\usepackage{lineno,hyperref}
\usepackage{amsmath, amssymb, amsfonts, wasysym, tipa, textcomp, amsbsy, graphics,subcaption, graphicx, setspace, fancyhdr, MnSymbol, mathrsfs , color,placeins, lipsum,xfrac,amsthm,listings, multirow}
\usepackage{booktabs}
\usepackage[utf8]{inputenc}
\usepackage[english]{babel}
\usepackage[labelfont=bf]{caption}
\usepackage{enumerate}
\definecolor{mygreen}{RGB}{28,172,0} 
\definecolor{mylilas}{RGB}{170,55,241}

\usepackage{float}
\restylefloat{table} 
\usepackage{tkz-graph}
\usetikzlibrary{arrows}
\usepackage{pinlabel}
\usepackage{float}
\floatstyle{plaintop}
\restylefloat{table}

\modulolinenumbers[250]

\begin{document}

\begin{frontmatter}

\title{Deep multi-modal networks for book genre classification based on its cover}

\author{Chandra Kundu, Lukun Zheng} 
\address{1906 College Heights Blvd, Bowling Green, KY 42101}

\author[mymainaddress]{Western Kentucky University} 

\cortext[mycorrespondingauthor]{Corresponding Author: Lukun Zheng}
\ead{lukun.zheng@wku.edu}


\begin{abstract}
Book covers are usually the very first impression to its readers and they often convey important information about the content of the book. Book genre classification based on its cover would be utterly beneficial to many modern retrieval systems, considering that the complete digitization of books is an extremely expensive task. At the same time, it is also an extremely challenging task due to the following reasons: First, there exists a wide variety of book genres, many of which are not concretely defined. Second, book covers, as graphic designs, vary in many different ways such as colors, styles, textual information, etc, even for books of the same genre. Third, book cover designs may vary due to many external factors such as country, culture, target reader populations, etc. With the growing competitiveness in the book industry, the book cover designers and typographers push the cover designs to its limit in the hope of attracting sales. The cover-based book classification systems become a particularly exciting research topic in recent years. In this paper, we propose a multi-modal deep learning framework to solve this problem. The contribution of this paper is four-fold. First, our method adds an extra modality by extracting texts automatically from the book covers. Second, image-based and text-based, state-of-the-art models are evaluated thoroughly for the task of book cover classification. Third, we develop an efficient  and salable multi-modal framework based on the images and texts shown on the covers only.  Fourth, a thorough analysis of the experimental results is given and future works to improve the performance is suggested. The results show that the multi-modal framework significantly outperforms the current state-of-the-art image-based models. However, more efforts and resources are needed for this classification task in order to reach a satisfactory level.
\end{abstract}


\end{frontmatter}

\linenumbers

\section{Introduction}
\begin{figure}[t]
		\centering
		\begin{subfigure}[t]{28mm}
    		\includegraphics[width=28mm, height=40mm]{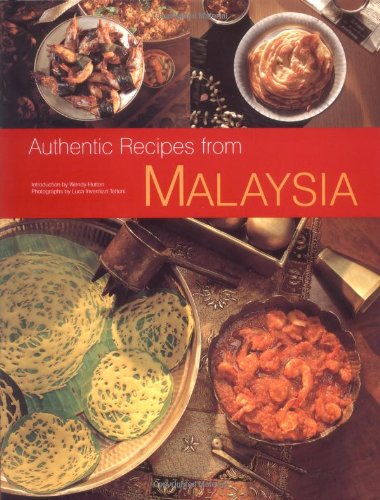}
    		\caption{}
		\end{subfigure}
		\begin{subfigure}[t]{28mm}
		    \includegraphics[width=28mm, height=40mm]{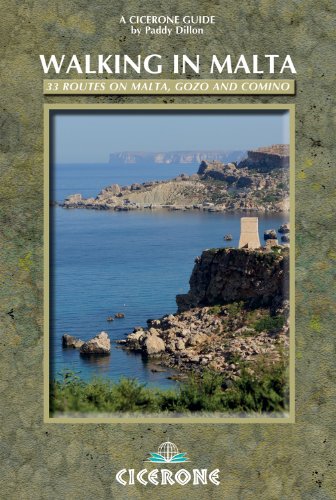}
		    \caption{}
		\end{subfigure}
		\begin{subfigure}[t]{28mm}
		    \includegraphics[width=28mm, height=40mm]{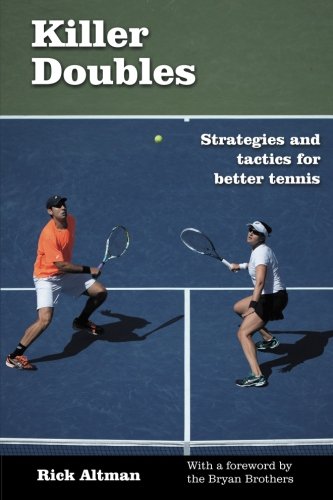}
		    \caption{}
		\end{subfigure}
		\begin{subfigure}[t]{28mm}
		    \includegraphics[width=28mm, height=40mm]{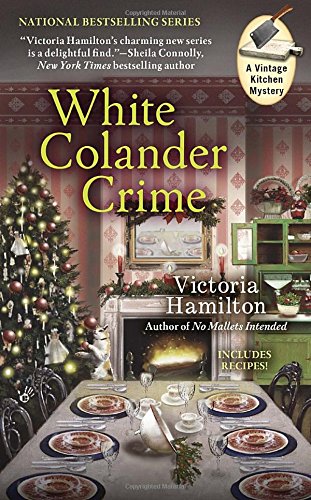}
		    \caption{}
		\end{subfigure}
		\caption{ Book covers conveying important information about the content of corresponding books. The genres are: (a) Cookbooks, Food \& Wine,
		(b) Travel, (c) Sports \& Outdoors and (d) Mystery, Thriller \& Suspense                       
		}
		\label{fig1:covers}
	\end{figure}

Books have been the one of the most important mediums for recording information and imparting knowledge in human history. Books can be classified into different categories based on their physical formats, contents, languages, and so on. In this paper, we focus on the task of book classification by its genre using the information provided just by the cover. Book covers are usually the very first impression to its readers and they often convey important information about the content of the book. Figure \ref{fig1:covers} presents some sample book covers. The information provided by a cover includes visual and textual information \cite{LSSRIUA2020}. For instance, in Figure 1(a), the background picture contains different food items and cookware which give the readers a visual impression about the book, while the texts shown on the cover states that it is a book about the ``authentic recipes from Malaysia". Both the visual and textual information are shown in the cover and they together indicate that its genre is ``Cookbooks, Food \& Wine". It is worth to mention that having only the visual information often makes the task extremely hard without textual information. For instance, in Figure 1 (d), without reading the texts on the cover, someone may classify the book as ``Cookbooks, Food \& Wine" as well solely based on the visual information we get from the cover that includes food items on a table in a dining room setting. Therefore, it is sometimes essential to consider both visual information and textual information extracted from the cover when we conduct book genre classification. The automatic classification of books based on only covers without human intervention would be utterly beneficial to many modern retrieval systems, considering that the complete digitization of books is an extremely expensive task.

The challenges of this task are the following. First, there exists a wide variety of book genres, many of which are not concretely defined. Second, book covers, as graphic designs, varies in many different ways such as colors, styles, textual information, etc, even for books of the same genre. Third, book cover designs may vary due to many external factors such as country, culture, target reader populations, etc \cite{Iwa2016}. To overcome these difficulties, we present a deep learning framework involving two moralities: one for visual information and the other for textual information extracted from the covers. 

Recently, deep learning approaches have reached high performances across a wide variety of problems \cite{Sud2016, Zha2019, Wan2019,ZZXW2019, LSZS2020,LKBSCGS2017,MY2017}. In particular, some deep convolutional neural networks can achieve a satisfactory level of performance on many visual recognition and categorization tasks, exceeding human performances. One of the most attractive qualities of these techniques is that they can perform well without any external hand-designed resources or task-specific feature engineering.  The theoretical foundations of deep learning are well rooted in the classical neural network (NN) literature. It involves many hidden neurons and layers as an architectural advantage in addition to the input and output layers \cite{Rav2016}. A deep convolutional neural network is universal, meaning that it can be used to approximate any continuous function to an arbitrary accuracy when the depth of the neural network is large enough \cite{Z2020}.

The main contributions of this paper are fourfold:
\begin{enumerate}[(1)]
\item Texts are extracted from the book covers in the benchmark dataset developed by Iwana et al. \cite{Iwa2016} using Google Cloud Vision API. It is worth to mention that the texts shown on the covers often contains the title, names of the author(s), publisher information, etc. These texts are directly extracted from the covers, which resembles the way people obtain both the visual and textual information from the book cover. 
\item State-of-the-art models are evaluated thoroughly for the task of book cover classification. These include image-only models and text-only models. 
\item  Two multi-modal models based on both images and texts directly extracted from the covers are proposed. Two different fusion methods to combine the information from both modalities are carefully developed and thoroughly analyzed. One is the simple concatenation and the other the deep canonical correlation analysis (Deep CCA).
\item A comprehensive analysis of the difficulties in this book cover classification task is provided and possible suggestions are made for future work to improve the performance. 
\end{enumerate}

The rest of the paper is structured as follows. Section 2 presents related works about book cover classification. Section 3 elaborates on the details of the proposed multi-modal architectures. In section 4, we discuss the experimental results. The last section concludes the paper and discusses future work. 
\section{Related Works}
In recent years, there has been an increasing interest in automated genre classification based on images by leveraging the strength of the deep neural network. Traditional machine learning methods often focus on feature engineering which extracts features using domain knowledge which are then used in the designed learning algorithm, while deep learning attempts to automatically learn features in large datasets through adjusting internal parameters using a backpropagation algorithm \cite{LBH2015}. There are multiple attempts to classify movie genre based on its poster with deep neural network \cite{CG2017,GS2019,KPS2020,PI2017}. Convolutional neural
networks have been used to categorize the genre of paintings and artworks \cite{CG2016,CLG2018,GH2020,ZGZ2020}. Similar works have been done in music genre classification as well \cite{KKS2016,OBNS2018,WHSW2019,YFWYL2020,YLLQLF2020}.

In book genre classification, Chiang et al. \cite{CGW2015} is the first attempt aimed to tackle this particular problem to the best of our knowledge. They implemented transfer learning with convolutional neural networks on the cover image along with natural language processing on the title text. A data set consisting of 6000 book covers from five genres obtained from OpenLibrary.org were utilized for their study. Iwana et al. \cite{Iwa2016} attempted to conduct book genre classification using only the visual clues provided by its cover. To solve this task, they created a large dataset consisting of 57,000 samples from 30 genres and adapted AlexNet \cite{KSH2012} pre-trained on ImageNet \cite{DDSLLF2009}. They achieved an overall accuracy rate of 24.7\% and 40.3\% for Top 1 and Top 3 predictions respectively.  Buczkowski et al. \cite{BSK2018} created another dataset consisting of 160k book covers crawled from GoodReads.com from over 500 genres, from which they picked the top 13 most popular genres and grouped all the remaining books under a 14th class called ``Others".  The authors used two different convolutional neural networks to predict book genres. One is relatively simple and shallow, with three convolutional layers, each followed directly by a $2\times 2$  max-pooling layer with non-overlapping windows. The other one adopted a more sophisticated architecture similar to the VGG network \cite{SZ2014}, with blocks of consecutive convolutional layers together with dropout layers. The authors achieved an accuracy
of 67\% using the more complex network and an accuracy of 74\% using the simpler network.  It is worth to note that the accuracy measure used in this study is not the traditional multi-class classification accuracy but a weighted score calculated based on the top 3 predicted genres. In \cite{BRVS2019}, the authors utilized a multinomial logistic regression model to classify book genres based on extracted image features and title features. Their approach consists of three stages: image feature extraction using the Xception model \cite{C2017}, title feature extraction using the GloVe Model \cite{PSM2014}, and classification based on the combined extracted features. Their study was based on a dataset consisting of 6076 samples from Amazon.com belonging to five genres. The authors achieved an overall accuracy of 87\% when they used the combined image and title features. Lucieri et al. \cite{LSSRIUA2020} aimed to benchmark deep learning models for classification of book covers. The authors used the same dataset introduced by Iwana et al. \cite{Iwa2016}. They provided a detailed evaluation of the state-of-the-art classification models for the task of book cover classification in an attempt to establish a benchmark on this problem. They employed the most powerful image recognition models such as NASNet, Inception ResNet v2, ResNet-50, etc. They provided a thorough analysis of the dataset and proposed a cleansed dataset by removing the book in the genre \emph{Reference} and merging the genre \emph{Christian
Books \& Bibles} with the genre \emph{Religion \& Spirituality}, by which they obtained a 28-category subset consisting of 55,100 samples. The authors also incorporated the title information available from the dataset and the text-image model yields the highest accuracy of 55.7\%. 

\section{Methodology: Multimodal Neural Networks}
\subsection{Data Set Preparation}
In this study, we used the BookCover30 dataset provided by Iwana et al. \cite{Iwa2016}, which consists of 57,000 book cover images divided equally into 30 genres, each genre with 1900 samples. We split the dataset into three parts: 80\% training samples, 10\% validation samples, and 10\% test samples.  All the images were then re-sized to 224 px by 224 px so that they have the same dimension. In order to obtain the textual information shown on the book covers, We apply Google Cloud Vision API to extract the texts from the cover images and then the extracted texts were encoded using a standand tokenization in which each word in the text is tokenized and then assigned a unique integer for representation. In this study we only encode words that appear at least 5 times in the whole dataset to avoid noises caused by insignificant words for classification. In other words, we extracted visual information and textual information from the covers. It is worth to note that the texts extracted from the book cover often includes the title of the book, the name(s) of the author(s), the publisher, and so on. This automatic extraction of both visual and textual information from the book cover resembles the process of human brains to obtain the visual and textual information from the book cover. It is worth noting that our approach differs itself from some of the existing works \cite{LSSRIUA2020,BRVS2019,CGW2015} in that they only used the title information stored in the original dataset, while we use all the texts automatically extracted from the cover image which often include texts besides the title. We first introduce some baseline image-based and text-based models and then the multimodal learning architectures proposed in this paper

\subsection{Image-Based and Text-Based Models}
\paragraph{Image-based models} We use several image-based models as our baseline models: LeNet \cite{LBBH1998}, AlexNet \cite{KSH2012}, VGGNet-16 \cite{SZ2014}, MobileNet-V1 \cite{HZCKWWA2017},, MobileNet-V2 \cite{SHZZC2018}, Inception-V2 \cite{SLVA2017},  and ResNet-50\cite{HSS2018}. These models were pre-trained on large datasets and achieved satisfactory performance on other similar image classification tasks. In this paper, we utilize these pre-trained models through transfer learning technique using the pre-trained values of the parameters involved in the model. 

\paragraph{Text-based models} In addition to these image-based models, we also evaluated two text-based models for this task. One is recurrent neural networks (RNN) with Long Short-Term Memory (LSTM) \cite{HS1997}. It allows the network to accumulate past information and thus be able to learn the long-term dependencies which are very common in textual data. The other one is Universal Sentence Encoder \cite{CYKHL2018}. It is a pre-trained sentence embedding and designed to leverages the encoder from Transformer which is a multi-attention head that helps the model ``attend" to the relevant information.

\begin{figure}[t]
\centering
\includegraphics[scale=.38]{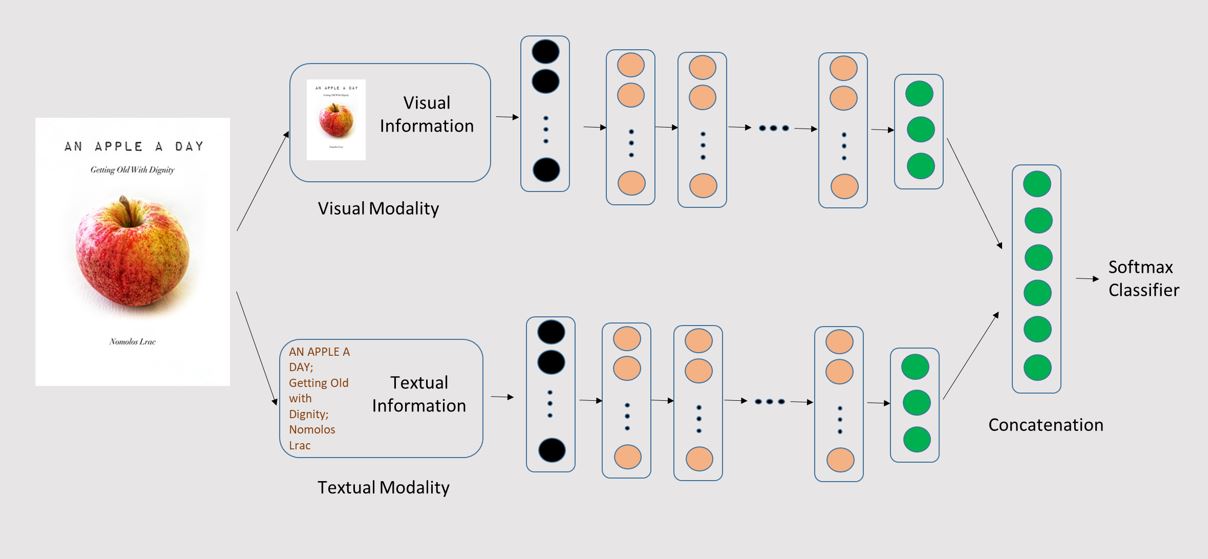}
\caption{The proposed multi-modal model with simple concatenation at a higher layer including deep networks (input layer, hidden layers, and output layer), concatenation layer and classifier softmax .}
\label{fig2:simple concatenation}
\end{figure}
\subsection{The Multi-Modal Model with Simple Concatenation}
The first proposed model is a multi-modal model including a visual modality and a textual modality with simple concatenation at a higher layer, which is shown in Figure \ref{fig2:simple concatenation}. The model contains three parts: deep networks (input layer, hidden layers, and output layer), feature concatenation and softmax classifier. The concatenation occurs at a higher layer instead of the input layer since concatenation at the input layer often causes 1) intractable training effort; 2) over-fitting due to pre-maturely learned features from both modalities; and 3) failure to learn implicit associations between modalities with different underlying features \cite{TYJ2018}. This model first learns the two modalities separately with two different flows, and then concatenate their features at a higher layer. It proves to be effective when two modalities have different basic representations, such as the visual data and the textual data from the book covers. We use the sparse categorical-cross-entropy loss to train the neural network.

\subsection{Multi-Modal Model with Deep CCA Concatenation}
Canonical correlation analysis (CCA) is a standard tool to find the linear projections between two different sets of random variables that are maximally correlated \cite{H1992}. Deep CCA or DCCA, as an extension of CCA, was proposed by Andrew et al. \cite{AABL2013} to learn representations of two views by passing them through multiple stacked layers of nonlinear transformations so that the two-most layers of two networks are maximally correlated. DCCA has been successfully used in a number of applications afterwards, such as acoustic features learning \cite{WALB2015}, RGB-D object recognition \cite{TYJ2018}, and emotion recognition \cite{LQZL2019}. Inspired by these successful applications, we consider how to utilize DCCA to tackle the task of book cover classification. 
\begin{figure}[t]
\centering
\includegraphics[scale=.38]{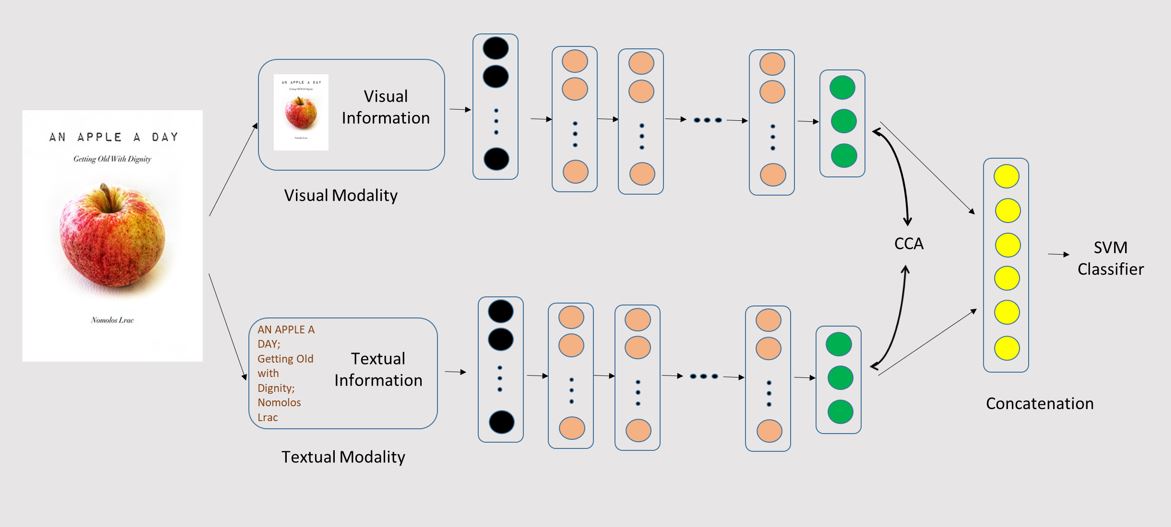}
\caption{The proposed multi-modal model with DCCA concatenation including deep networks (input layer, hidden layers, and output layer), Canonical Correlation Analysis concatenation layer and classifier SVM.}
\label{fig3:DCCA concatenation}
\end{figure}
Figure \ref{fig3:DCCA concatenation} shows the multi-modal model with Deep CCA concatenation, which includes three parts: deep networks (input layer, hidden layers, and output layer), Canonical Correlation Analysis concatenation layer and classifier SVM. This technique allows the multi-modal model to learn correlated features between the visual modality and the textual modality. Back propagation is utilized to these deep networks for feature extraction so that the correlation between the top layers of them is maximized. Then the extracted features from these two modalities are concatenated at a higher layer, followed by a support vector machine (SVM) classifier. 

\section{Experimental Results}
\subsection{Implementation}
\begin{itemize}
\item[(1)] \emph{Image-based models}. The image-based models mentioned in the previous section, we utilized the pre-trained models on ImageNet dataset and added a flatten layer of 1024 hidden units, and an output layer of 30 units on top of every pre-trained models. The softmax activation function was used for the output layer. We used sparse-categorical-cross-entropy loss with Adam optimizer with a default learning rate of 0.001 for all these models. 
\item[(2)] \emph{Text-based models}. We used a simple vanilla LSTM with 256 memory units for the text-based RNN model. We used the pre-trained Universal Sentence Encoder model \cite{CYKHL2018}. The softmax activation function was used for the output layer. We used sparse-categorical-cross-entropy loss with Adam optimizer with a default learning rate of 0.001 in both models.. 
\item[(3)] \emph{Multi-modal models}. We chose the best image-based model (ResNet-50) and use it in the visual modality and chose the best text-based model (Universal Sentence Encoder) and use it in the textual modality. For the multi-modal model with simple concatenation, we freeze the chosen pre-trained models and extract features using the ReLU activation function. Then we concatenate these features from both modalities and feed them into a final fully connected layer with a softmax activation function. For the multi-modal model with DCCA concatenation, we use similar procedures except that the CCA loss was utilized as the loss function and an SVM classifier was trained on the features from the concatenated layers. 
\item[(4)] \emph{Experimental setup}. All the experiments are executed with Python in Google Colab using a Nvidia Tesla K80 GPU. We use Keras (an open-source deep learning library) with the TensorFlow backend. 
\end{itemize}
\subsection{Results}

\begin{table}[t]
\centering
\caption{Accuracy comparison of the models including the image-based models, text-based models, and multi-modal models on the test set}
\begin{tabular}{p{0.58\textwidth}llp{0.25\textwidth}lp{0.25\textwidth}}
\hline\hline 
&\multicolumn{2}{c}{Accuracy}\\[0.3ex]\cmidrule{2-3} 
Models&Top 1(\%)  &Top 3 (\%)\\  [0.5ex] \hline
\emph{Image-based models}&&\\[1.2ex]
LetNet \cite{Iwa2016} & 13.5 &24.7\\
AlexNet \cite{Iwa2016}& 24.7 &40.3\\
VGGNet-16 &25.6 & 45.6\\
MobileNet-V1 &27.2 &46.4\\
MobileNet-V2 &23.6 &42.0\\
Inception-V2 &26.2 &45.6\\
\textbf{ResNet-50} &\textbf{29.6} &\textbf{49.0} \\ [0.5ex]\hline
\emph{Text-based models}&&\\[1.2ex]
RNN-LSTM &41.5 &61.6\\
\textbf{Universal Sentence Encoder} &\textbf{52.6} &\textbf{73.7}\\ [0.5ex]\hline
\emph{Multi-modal models} &&  \\[1.2ex]
\textbf{Simple concatenation} &\textbf{56.1} &\textbf{77.7}\\
DCCA concatenation & 48.9 &72.3\\\hline\hline
\end{tabular}
\label{tab: accuracy table}
\end{table}

Table \ref{tab: accuracy table} shows the classification accuracy comparison among the models including  the image-based models, text-based models, and multi-modal models on the test set. We included the accuracy rates of LeNet and AlexNet presented in Iwana et al. \cite{Iwa2016} as baselines. Among the image-based models, ResNet-50 achieved the best performance with a top-1 accuracy of 19.6\% and a top-3 accuracy of 49.0\%, a 6\% gain in the top-1 and 9\% gain in top-3 accuracy over the baseline. The two text-based models had much better results. The RNN-LSTM model achieved a top-1 accuracy of 41.5\% and a top-3 accuracy of 61.6\%. The Universal Sentence Encoder model achieved a top-1 accuracy of 52.6\% and a top-3 accuracy of 73.7, a 28\% gain in top-1 accuracy and a 33\% gain in top-3 accuracy over the baseline. Finally, the multi-modal models had the best performance. The DCCA concatenation model achieved a top-1 accuracy of 48.9\% and a top-3 accuracy of 72.3\%. The simple concatenation model achieved a top-1 accuracy of 56.1\% and a top-3 accuracy of 77.7\%, a 30\% gain in top-1 accuracy and a 33\% gain on top-3 accuracy over the baseline. 

\begin{table}[H]
\centering
\caption{Accuracy comparison of between the baseline and the proposed multi-modal model with simple concatenation for individual genres. Here Simple-C represents the multi-modal model with simple concatenation and DCCA-C represents the multi-modal model with DCCA concatenation}
\def\arraystretch{0.8}
\begin{tabular}{lcclccl}
\hline\hline 
&\multicolumn{2}{c}{Baseline} &&\multicolumn{2}{c}{Multi-Modal Models}\\[0.3ex] \cmidrule{2-3}  \cmidrule{5-6} 
Genre &LetNet  &AlexNet &&Simple-C  &DCCA-C\\  [0.5ex]
\hline 
Arts \& Photography &5.8&12.1 && 43.8 &39.5  \\
Biographies \& Memoirs &5.3&13.2  && 38.8 & 33.1 \\
Business \& Money&10.0 &12.6  && 60.6 &48.7  \\
Calendars&18.9 &47.9  && 85.4 & 80.6 \\
Children's Books&24.7 &42.1  && 51.5 & 52.0 \\
Comics \& Graphic Novels& 15.8&47.4 && 68.8 &66.7  \\
Computers \& Technology&29.5 &44.7  && 76.9 & 69.7 \\
Cookbooks, Food \& Wine& 14.2&43.7  && 88.1 &83.7  \\
Crafts, Hobbies \& Home&7.4 &17.4  && 66.1 &46.2  \\
Christian Books \& Bibles&8.4 &7.4  && 60.4 & 54.3 \\
Engineering \& Transportation&10.0 & 20.07&& 61.3 &50.3  \\
Health, Fitness \& Dieting&4.2 &12.6  && 52.6 &39.9  \\
History&6.3 &12.6  && 50.6 & 44.1 \\
Humor \& Entertainment&5.3 &10.5  && 38.1 &22.0 \\
Law& 14.7&25.3  && 68.8 & 47.2 \\
Literature \& Fiction&3.2 &11.1  && 27.5 & 28.3 \\
Medical Books&12.6 &19.5  && 73.8 &55.9  \\
Mystery, Thriller \& Suspense&23.7 &34.2  && 68.2 &60.7  \\
Parenting \& Relationships&14.7 &24.2  && 52.1 & 49.0 \\
Politics \& Social Sciences&3.7 & 6.8 && 35.4 &22.3  \\
Reference& 13.2&20.0  && 41.7 &34.5  \\
Religion \& Spirituality&8.4 &16.3  && 55.6 &44.1  \\
Romance& 27.4&45.3  && 60.4 & 66.3 \\
Science \& Math&8.4 &14.2  && 60.0 & 40.2 \\
Science Fiction \& Fantasy&27.4 &35.8 && 50.5 & 48.3 \\
Self-Help&13.7 & 14.2 && 53.8 & 44.5 \\
Sports \& Outdoors&5.3 & 14.7 && 60.9 & 49.3 \\
Teen \& Young Adult&7.9 &12.1  && 14.6 &18.6  \\
Test Preparation& 47.9&68.9  &&81.3 & 76.9 \\
Travel&19.5 &33.2  && 54.7 & 50.5 \\\hline\hline
\end{tabular}
\label{tab: genre accuracy comparison}
\end{table}
Table \ref{tab: genre accuracy comparison} presents the classification accuracy comparison on the test data between the baseline models in \cite{Iwa2016} and the proposed multi-modal models  for individual genres. There are several observations from this table. First, In comparison, it can be seen that the proposed multi-modal models significantly outperform the baseline models for all individual genres. Second, for some of the genres, our proposed models, especially the simple concatenation model, achieved high accuracy rates. For instance, the simple concatenation model achieved an accuracy of 88.1\% in ``Cookbooks, Food \& Wine" and an accuracy of 85.4\% in ``Calendars". The DCCA concatenation model achieved an accuracy of 83.7\% in ``Cookbooks, Food \& Wine" and an accuracy of 80.6\% in ``Calendars". Third, there are genres for which even our proposed models perform poorly on this task. For instance, the simple concatenation model and the DCCA concatenation model have accuracy rates of 14.6\%, and 18.6\% for ``Teen \& Young Adult", respectively. Finally, the simple concatenation model outperforms the DCCA concatenation model in all genres except ``Childcare's Books", ``Literature \& Fiction", ``Romance",  and ``Teen \& Young Adult". DCCA model yields a 0.5\% gain over the simple concatenation model in ``Childcare's Books" a 0.8\% gain in ``Literature \& Fiction", a 5.9\% gain in ``Romance", and a 4\% gain in ``Teen \& Young Adult"

\begin{figure}[ht]
\centering
\includegraphics[scale=.4]{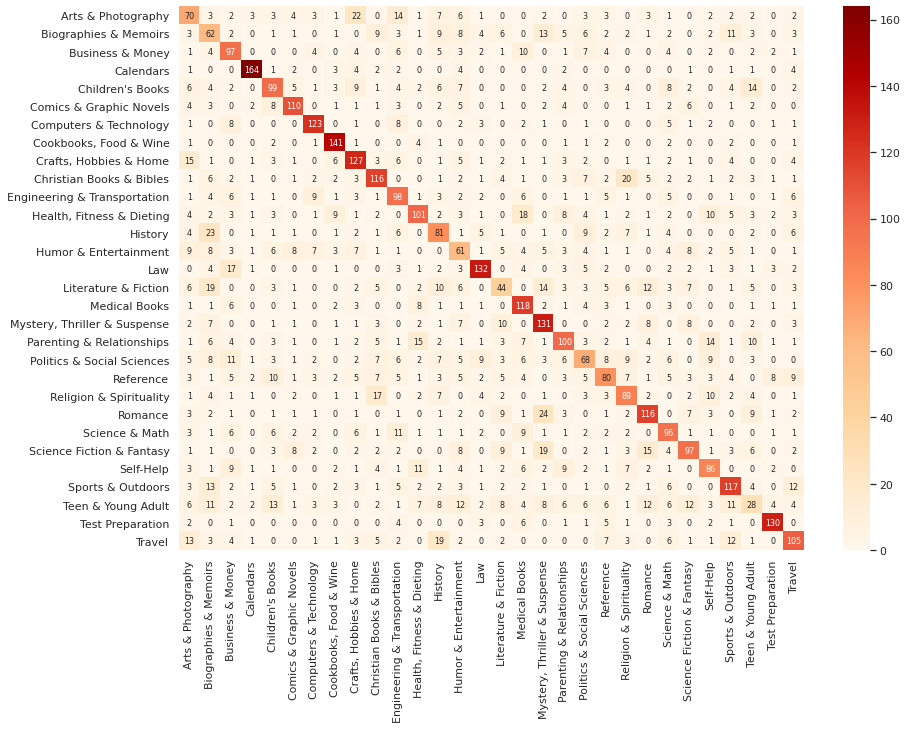}
\caption{Confusion matrix of the simple concatenation model on the test data. The horizontal axis represents the actual genres and the vertical axis represents the predicted genres.}
\label{fig: confusion-matrix}
\end{figure}

Figure \ref{fig: confusion-matrix} presents the confusion matrix of the simple concatenation model on the test data, with the horizontal axis representing the actual genres and the vertical axis representing the predicted genres. The diagonal entries represent the number of covers of each genre in the test data that are classified correctly by the simple concatenation model. For instance, the entry ``164" at the (4,4) position of the matrix tells that 164 book covers from ``Calendar" are classified correctly as ``Calendar", resulting in a top-1 accuracy of 164/192=85.4\%, where 192 is the total number of covers with the genre ``Calendar" from the test data. The off-diagonal entries provide detailed information about the numbers of misclassifications among each genre. For instance, the entry ``2" in the (6, 4) position of the matrix tells that 2 book covers in ``Calendar" were misclassified as ``Comics \& Graphic Novels". One observation is that books in ``Teen \& Young Adult" were misclassified to almost all other genres, which makes sense since many books of different genres are often written targeting on young adults. Another striking misclassification is the one of ``Biographies \& Memoirs" together with ``History", which is again understandable, as many books in ``Biographies \& Memoirs" are about historical figures.  Other significant misclassifications can also be observed. For instance, ``Christian Books \& Bible" books were sometimes misclassified as ``Religion \& Spirituality", since Christian books are a subset of religion books. The reason why Christian books were singled out as a separate genre might be that it addresses the majority of the American customers due to the fact that the data were collected through  the \emph{Amazon.com} page in the United States of America.

\subsection{Discussion}
Even though the proposed multi-modal models have much better performance than the baseline models for all the individual genres as shown in Table \ref{tab: genre accuracy comparison}, their performances in some of the genres such as ``Teen \& Young Adult" are far from satisfactory. In this section, we summarize the reasons why most of the state-of-the-art models fail in this task. 

\begin{figure}[t]
		\centering
		\begin{subfigure}[t]{36mm}
		    \includegraphics[width=36mm, height=50mm]{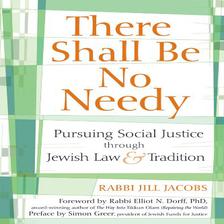}
		    \caption{}
		\end{subfigure}
		\begin{subfigure}[t]{36mm}
		    \includegraphics[width=36mm, height=50mm]{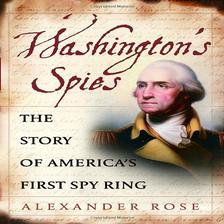}
		    \caption{}
		\end{subfigure}
		\begin{subfigure}[t]{36mm}
		    \includegraphics[width=36mm, height=50mm]{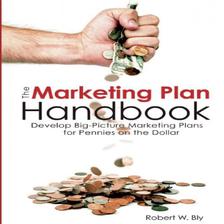}
		    \caption{}
		\end{subfigure}
		\caption{The randomization of genre assignments. (a) Religion \& Spirituality; (b) Biographies \& Memoirs; (c) Self-help. 
		}
		\label{fig4:problem-genre randomization}
	\end{figure}

First, there is a major issue in the dataset in Iwana et al. \cite{Iwa2016}. The dataset was
collected from the book cover images and genres listed by Amazon.com as the top
categories under ``Books". In the source data, many of the books belong to multiple
genres. However, during the prepossessing, the authors randomly selected one genre for each book cover for simplicity. This randomization of genre assignment raise concerns on the validity of the 
dataset and poses great complexity to the problem. For instance, the book cover (a) in Figure \ref{fig4:problem-genre randomization} is mainly a law book which touches upon religion and traditional Jewish cultures and hence belonged to two genres: ``Law", and "Religion \& Spirituality" in the source data. However, in the randomization process, the latter genre was assigned as its genre. The book cover (b) in Figure \ref{fig4:problem-genre randomization} belonged to both ``History" and ``Biographies \& Memoirs" in the source data but was assigned to the latter one during randomization. The book cover (c) in Figure \ref{fig4:problem-genre randomization} belonged to ``Business \& Money" and ``Self-Help" in the source data and was assigned to ``Self-Help". 

\begin{figure}[t]
		\centering
		\begin{subfigure}[t]{28mm}
    		\includegraphics[width=28mm, height=40mm]{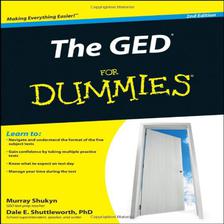}
    		\caption{}
		\end{subfigure}
		\begin{subfigure}[t]{28mm}
		    \includegraphics[width=28mm, height=40mm]{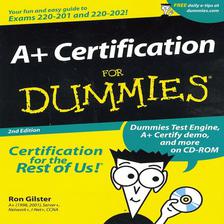}
		    \caption{}
		\end{subfigure}
		\begin{subfigure}[t]{28mm}
		    \includegraphics[width=28mm, height=40mm]{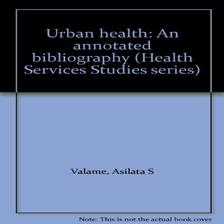}
		    \caption{}
		\end{subfigure}
		\begin{subfigure}[t]{28mm}
		    \includegraphics[width=28mm, height=40mm]{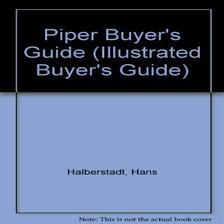}
		    \caption{}
		\end{subfigure}
		\caption{Low inter-genre variance. (a) genre: Test Preparation; (b) genre: Computers \& Technology; (c) genre: Reference; (d) genre: Engineering \& Transportation. 
		}
		\label{fig5:low-inter-variance}
	\end{figure}

\begin{figure}[t]
		\centering
		\begin{subfigure}[t]{28mm}
    		\includegraphics[width=28mm, height=40mm]{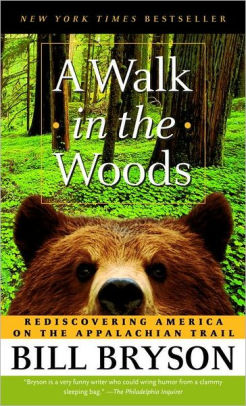}
    		\caption{}
		\end{subfigure}
		\begin{subfigure}[t]{28mm}
		    \includegraphics[width=28mm, height=40mm]{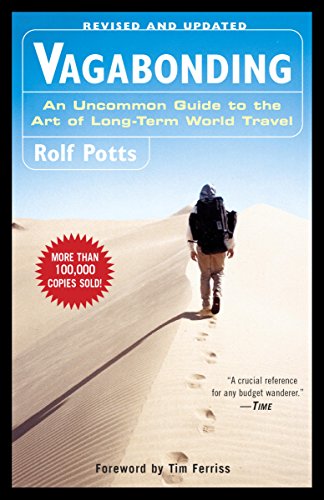}
		    \caption{}
		\end{subfigure}
		\begin{subfigure}[t]{28mm}
		    \includegraphics[width=28mm, height=40mm]{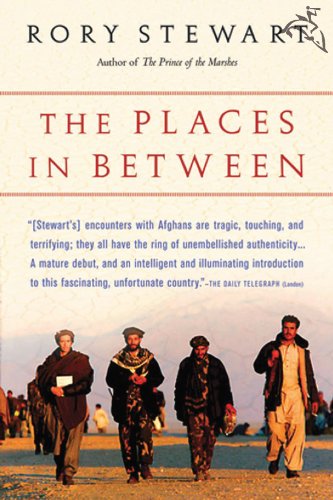}
		    \caption{}
		\end{subfigure}
		\begin{subfigure}[t]{28mm}
		    \includegraphics[width=28mm, height=40mm]{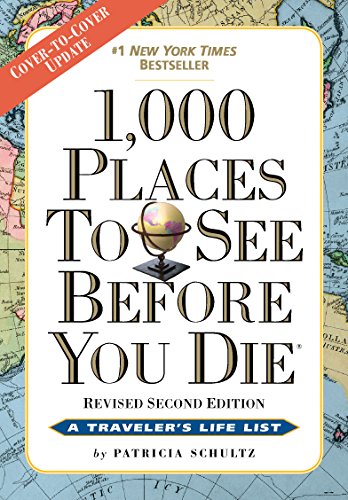}
		    \caption{}
		\end{subfigure}
		\caption{High intra-genre variance. All of these books belong to the same genre ``Travel" but their covers are very different.  
		}
		\label{fig6:high-intra-variance}
	\end{figure}

Second, the dataset exhibits low inter-genre variance and high intra-genre variance in some genres \cite{LSSRIUA2020}. Low inter-genre variance refers to the fact that book covers of different genres look very similar. For instance, Figures \ref{fig5:low-inter-variance} (a) and \ref{fig5:low-inter-variance} (b) are very similar to each other but they belong to different genres. Similarly, Figures \ref{fig5:low-inter-variance} (c) and \ref{fig5:low-inter-variance} (d) are very similar to each other but they belong to different genres. High intra-genre variance refers to the fact that book covers of the same genre are very different from each other. For instance, all the books in Figure \ref{fig6:high-intra-variance} are from the genre ``Travel" but look very different from each other. The existence of these variances adds complexity to the task and makes it extremely difficult to deal with. 

\begin{figure}[t]
		\centering
		\begin{subfigure}[t]{28mm}
    		\includegraphics[width=28mm, height=40mm]{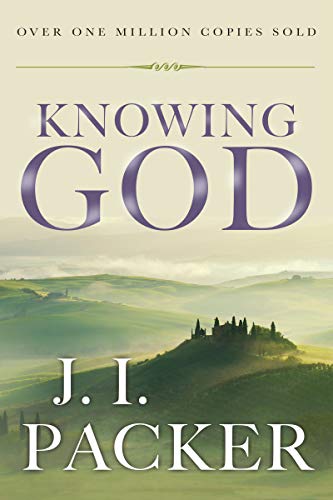}
    		\caption{}
		\end{subfigure}
		\begin{subfigure}[t]{28mm}
		    \includegraphics[width=28mm, height=40mm]{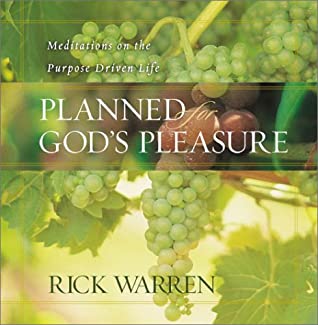}
		    \caption{}
		\end{subfigure}
		\begin{subfigure}[t]{28mm}
		    \includegraphics[width=28mm, height=40mm]{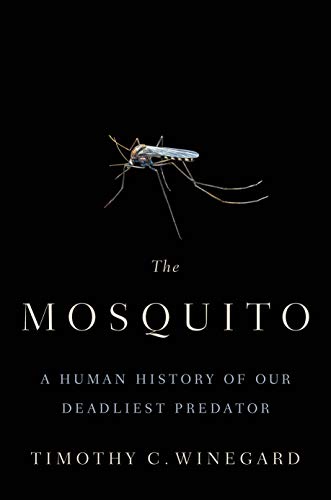}
		    \caption{}
		\end{subfigure}
		\begin{subfigure}[t]{28mm}
		    \includegraphics[width=28mm, height=40mm]{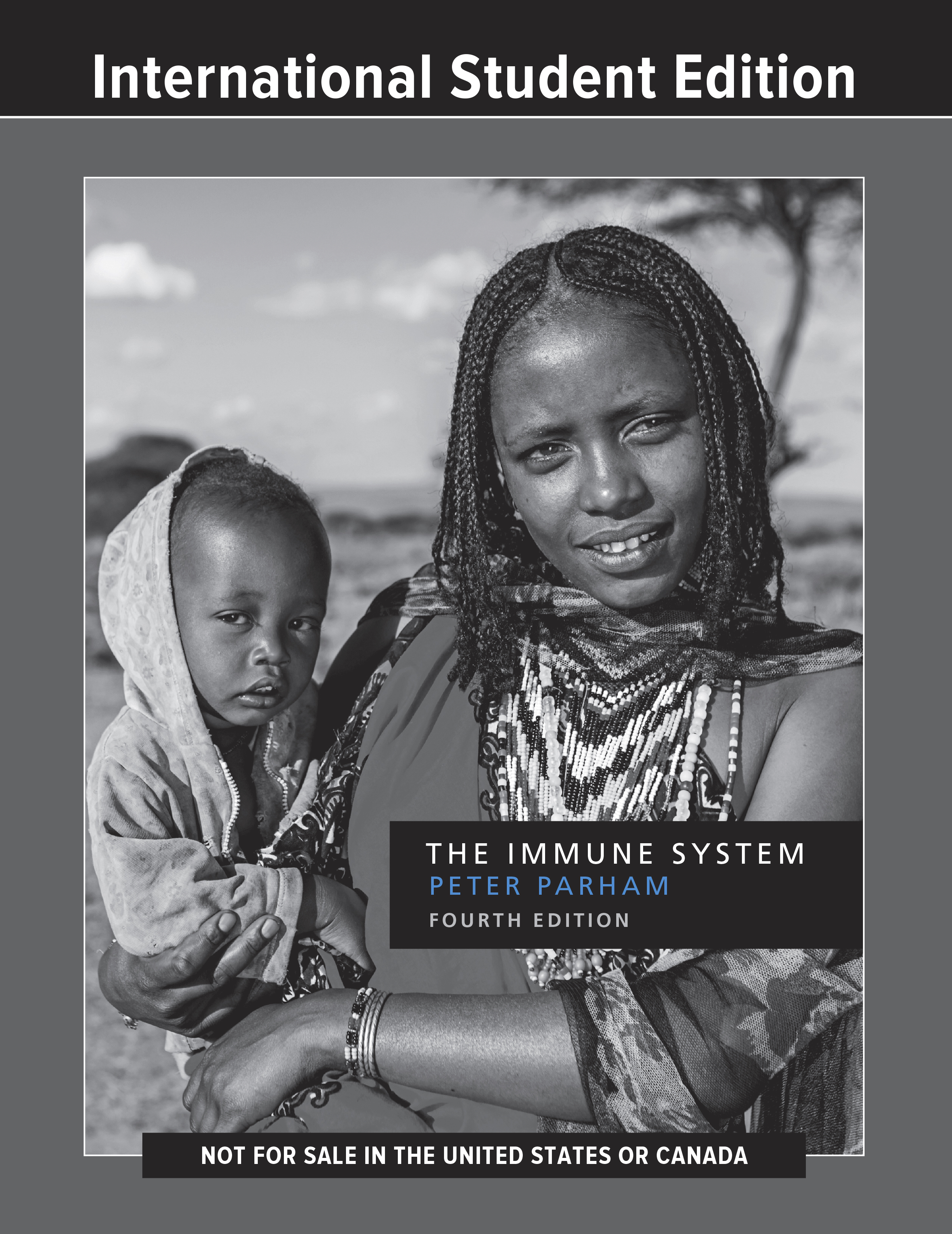}
		    \caption{}
		\end{subfigure}
		\caption{Inconsistent textual and visual information from the book cover. (a) texts suggest ``Christian Books \& Bible", while image indicates ``Travel"; (b) texts suggest ``Christian Books \& Bible", while image suggests ``Cookbooks, Food \& Wine"; (c) Texts suggest ``History", while image suggests ``Arts \& Photography"; (d) Texts suggest ``Medical Books", while image suggests ``Parenting \& Relationships".
		}
		\label{fig7:inconsistant-tx-image}
	\end{figure}	
	Another factor that makes it difficult for prediction in this task is that the texts and image shown on the same book cover may convey inconsistent information regarding the genre of the book. For instance, in Figure \ref{fig7:inconsistant-tx-image} (a),  the texts suggests ``Christian Books \& Bible", while image indicates ``Travel"; in Figure \ref{fig7:inconsistant-tx-image} (b),  the texts suggest ``Christian Books \& Bible", while image indicates ``Cookbooks, Food \& Wine";  in Figure \ref{fig7:inconsistant-tx-image} (c),  the texts suggest ``History", while image indicates ``Arts \& Photography"; in Figure \ref{fig7:inconsistant-tx-image} (d),  the texts suggest ``Medical Books", while image indicates ``Parenting \& Relationships". The existence of such covers with inconsistent textual and visual information have a negative impact on the performance of the DCCA concatenation model since DCCA would suffer if the information from these two modalities provide inconsistent information regarding the classification task. 
	
\section{Conclusion}
In this paper, we proposed two multi-modal models: one with simple concatenation and the other with DCCA concatenation, for the task of book genre classification solely based on its cover. In addition, we evaluated several state-of-the-art image-based models and text-based models. By comparison, text-based models perform better in general than image-based models and the proposed multi-modal model with simple concatenation outperforms all other models. Based on the results from our experiments, the simple concatenation model has a top-1 accuracy of 56.1\%, more than doubling that of the baseline image-based model AlexNet introduced in \cite{Iwa2016}.  The DCCA concatenation model has a top-1 accuracy of 48.9\%, which is less than that of the simple concatenation model. A reasonable explanation for this poorer performance is that the textual and visual information in some book covers are inconsistent regarding the task of genre classification. We also outlined several drawbacks about the dataset which may cause problems for this task. In order to solve this task to a better level of performance, more efforts are needed for the creation of better book cover dataset and the development of more sophisticated models. 

\vspace{.2 cm}  

\noindent \textbf{Acknowledgements} We would like to thank Dr. Mark Robinson, Dr. Ngoc Nguyen, and Dr. Qi Li from Western Kentucky University for their valuable comments and suggestions to improve this work.


\begin{thebibliography}{99}
\bibitem{AABL2013} Andrew, G., Arora, R., Bilmes, J., \& Livescu, K. (2013). Deep canonical correlation analysis. In International conference on machine learning (pp. 1247-1255).

\bibitem{BRVS2019} Biradar, G. R., Raagini, J. M., Varier, A., \& Sudhir, M. (2019). Classification of Book Genres using Book Cover and Title. In 2019 IEEE International Conference on Intelligent Systems and Green Technology (ICISGT) (pp. 72-723). IEEE.

\bibitem{BSK2018} Buczkowski, P., Sobkowicz, A., \& Kozlowski, M. (2018). Deep Learning Approaches towards Book Covers Classification. In ICPRAM (pp. 309-316).



\bibitem{CYKHL2018} Cer, D., Yang, Y., Kong, S. Y., Hua, N., Limtiaco, N., John, R. S., ... \& Sung, Y. H. (2018). Universal sentence encoder. arXiv preprint arXiv:1803.11175.

\bibitem{CG2016} Cetinic, E., \& Grgic, S. (2016). Genre classification of paintings. In 2016 International Symposium ELMAR (pp. 201-204). IEEE.

\bibitem{CLG2018}Cetinic, E., Lipic, T., \& Grgic, S. (2018). Fine-tuning convolutional neural networks for fine art classification. Expert Systems with Applications, 114, 107-118.



\bibitem{CGW2015} Chiang, H., Ge, Y., \& Wu, C. (2015). Classification of Book Genres By Cover and Title.

\bibitem{C2017} Chollet, F. (2017). Xception: Deep learning with depthwise separable convolutions. In Proceedings of the IEEE conference on computer vision and pattern recognition (pp. 1251-1258).

\bibitem{CG2017} Chu, W. T., \& Guo, H. J. (2017, October). Movie genre classification based on poster images with deep neural networks. In Proceedings of the Workshop on Multimodal Understanding of Social, Affective and Subjective Attributes (pp. 39-45).

\bibitem{DDSLLF2009} Deng, J., Dong, W., Socher, R., Li, L. J., Li, K., \& Fei-Fei, L. (2009). Imagenet: A large-scale hierarchical image database. In 2009 IEEE conference on computer vision and pattern recognition (pp. 248-255). Ieee.

\bibitem{GS2019}Gozuacik, N., \& Sakar, C. O. (2019). Turkish Movie Genre Classification from Poster Images using Convolutional Neural Networks. In 2019 11th International Conference on Electrical and Electronics Engineering (ELECO) (pp. 930-934). IEEE.

\bibitem{GH2020}Guo, B., \& Hao, P. (2020). Analysis of Artistic Styles in oil Painting using Deep-Learning Features. In 2020 IEEE International Conference on Multimedia \& Expo Workshops (ICMEW) (pp. 1-4). IEEE.

\bibitem{HS1997} Hochreiter, S., \& Schmidhuber, J. (1997). Long short-term memory. Neural computation, 9(8), 1735-1780.

\bibitem{H1992} Hotelling, H. (1992). Relations between two sets of variates. In Breakthroughs in statistics (pp. 162-190). Springer, New York, NY.

\bibitem{HZCKWWA2017}Howard, A. G., Zhu, M., Chen, B., Kalenichenko, D., Wang, W., Weyand, T., ... \& Adam, H. (2017). Mobilenets: Efficient convolutional neural networks for mobile vision applications. arXiv preprint arXiv:1704.04861.

\bibitem{HSS2018} Hu, J., Shen, L., \& Sun, G. (2018). Squeeze-and-excitation networks. In Proceedings of the IEEE conference on computer vision and pattern recognition (pp. 7132-7141).

\bibitem{Iwa2016} Iwana, B. K., Rizvi, S. T. R., Ahmed, S., Dengel, A., \& Uchida, S. (2016). Judging a book by its cover. arXiv preprint arXiv:1610.09204.

\bibitem{KKS2016}Kim, S., Kim, D., \& Suh, B. (2016). Music genre classification using multimodal deep learning. In Proceedings of HCI Korea (pp. 389-395).

\bibitem{KSH2012} Krizhevsky, A., Sutskever, I., \& Hinton, G. E. (2012). Imagenet classification with deep convolutional neural networks. In Advances in neural information processing systems (pp. 1097-1105).

\bibitem{KPS2020}Kundalia, K., Patel, Y., \& Shah, M. (2020). Multi-label movie genre detection from a movie poster using knowledge transfer learning. Augmented Human Research, 5(1), 11.

\bibitem{LBH2015}LeCun, Y., Bengio, Y., \& Hinton, G. (2015). Deep learning. nature, 521(7553), 436-444.

\bibitem{LBBH1998}LeCun, Y., Bottou, L., Bengio, Y., \& Haffner, P. (1998). Gradient-based learning applied to document recognition. Proceedings of the IEEE, 86(11), 2278-2324.

\bibitem{LKBSCGS2017} Litjens, G., Kooi, T., Bejnordi, B. E., Setio, A. A. A., Ciompi, F., Ghafoorian, M., ... \& Sánchez, C. I. (2017). A survey on deep learning in medical image analysis. Medical image analysis, 42, 60-88.

\bibitem{LQZL2019} Liu, W., Qiu, J. L., Zheng, W. L., \& Lu, B. L. (2019). Multimodal emotion recognition using deep canonical correlation analysis. arXiv preprint arXiv:1908.05349.

\bibitem{LSZS2020}Lotfollahi, M., Siavoshani, M. J., Zade, R. S. H., \& Saberian, M. (2020). Deep packet: A novel approach for encrypted traffic classification using deep learning. Soft Computing, 24(3), 1999-2012.

\bibitem{LSSRIUA2020}Lucieri, A., Sabir, H., Siddiqui, S. A., Rizvi, S. T. R., Iwana, B. K., Uchida, S., ... \& Ahmed, S. (2020). Benchmarking Deep Learning Models for Classification of Book Covers. SN Computer Science, 1, 1-16.

\bibitem{MY2017} Min, S., Lee, B., \& Yoon, S. (2017). Deep learning in bioinformatics. Briefings in bioinformatics, 18(5), 851-869.

\bibitem{OBNS2018}Oramas, S., Barbieri, F., Nieto, O., \& Serra, X. (2018). Multimodal deep learning for music genre classification. Transactions of the International Society for Music Information Retrieval. 2018; 1 (1): 4-21.

\bibitem{PSM2014}Pennington, J., Socher, R., \& Manning, C. D. (2014). Glove: Global vectors for word representation. In Proceedings of the 2014 conference on empirical methods in natural language processing (EMNLP) (pp. 1532-1543).

\bibitem{PI2017}Pobar, M.,\& Ivasic-Kos, M. (2017). Multi-label poster classification into genres using different problem transformation methods. In International conference on computer analysis of images and patterns (pp. 367-378). Springer, Cham.

\bibitem{Rav2016} Ravì, D., Wong, C., Deligianni, F., Berthelot, M., Andreu-Perez, J., Lo, B., \& Yang, G. Z. (2016). Deep learning for health informatics. IEEE journal of biomedical and health informatics, 21(1), 4-21.

\bibitem{SHZZC2018} Sandler, M., Howard, A., Zhu, M., Zhmoginov, A., \& Chen, L. C. (2018). Mobilenetv2: Inverted residuals and linear bottlenecks. In Proceedings of the IEEE conference on computer vision and pattern recognition (pp. 4510-4520).

\bibitem{SZ2014} Simonyan, K., \& Zisserman, A. (2014). Very deep convolutional networks for large-scale image recognition. arXiv preprint arXiv:1409.1556.

\bibitem{Sud2016} Sudholt, S., \& Fink, G. A. (2016, October). Phocnet: A deep convolutional neural network for word spotting in handwritten documents. In 2016 15th International Conference on Frontiers in Handwriting Recognition (ICFHR) (pp. 277-282). IEEE.

\bibitem{SLVA2017} Szegedy, C., Ioffe, S., Vanhoucke, V., \& Alemi, A. A. (2017, February). Inception-v4, inception-resnet and the impact of residual connections on learning. In Thirty-first AAAI conference on artificial intelligence.

\bibitem{TYJ2018} Tang, L., Yang, Z. X., \& Jia, K. (2018). Canonical correlation analysis regularization: an effective deep multiview learning baseline for RGB-D object recognition. IEEE Transactions on Cognitive and Developmental Systems, 11(1), 107-118.

\bibitem{WHSW2019} Wu, W., Han, F., Song, G., \& Wang, Z. (2019). Music Genre Classification Using Independent Recurrent Neural Network. In 2018 Chinese Automation Congress (CAC) (pp. 192-195). IEEE.
\bibitem{Wan2019} Wang, J., Chen, Y., Hao, S., Peng, X., \& Hu, L. (2019). Deep learning for sensor-based activity recognition: A survey. Pattern Recognition Letters, 119, 3-11.

\bibitem{WALB2015}Wang, W., Arora, R., Livescu, K., \& Bilmes, J. A. (2015). Unsupervised learning of acoustic features via deep canonical correlation analysis. In 2015 IEEE International Conference on Acoustics, Speech and Signal Processing (ICASSP) (pp. 4590-4594). IEEE.

\bibitem{YFWYL2020} Yang, R., Feng, L., Wang, H., Yao, J., \& Luo, S. (2020). Parallel Recurrent Convolutional Neural Networks-Based Music Genre Classification Method for Mobile Devices. IEEE Access, 8, 19629-19637.

\bibitem{YLLQLF2020}  Yu, Y., Luo, S., Liu, S., Qiao, H., Liu, Y., \& Feng, L. (2020). Deep attention based music genre classification. Neurocomputing, 372, 84-91.

\bibitem{ZGZ2020}Zeng, Y., Gong, Y., \& Zeng, X. (2020). Controllable digital restoration of ancient paintings using convolutional neural network and nearest neighbor. Pattern Recognition Letters, 133, 158-164.

\bibitem{Zha2019}Zhang, S., Yao, L., Sun, A., \& Tay, Y. (2019). Deep learning based recommender system: A survey and new perspectives. ACM Computing Surveys (CSUR), 52(1), 1-38.

\bibitem{ZZXW2019}Zhao, Z. Q., Zheng, P., Xu, S. T., \& Wu, X. (2019). Object detection with deep learning: A review. IEEE transactions on neural networks and learning systems, 30(11), 3212-3232.

\bibitem{Z2020} Zhou, D. X. (2020). Universality of deep convolutional neural networks. Applied and computational harmonic analysis, 48(2), 787-794.
\end{thebibliography}
\end{document}